\documentclass[journal]{IEEEtran}
\usepackage{amsmath,amsfonts}
\usepackage{todonotes}
\usepackage{algorithmic}
\usepackage{algorithm}
\usepackage{array}
\usepackage{microtype}
\usepackage{textcomp}
\usepackage{stfloats}
\usepackage{url}
\usepackage{verbatim}
\usepackage{graphicx}
\usepackage{cite}
\usepackage{orcidlink}
\usepackage{longtable}
\usepackage{threeparttable}
\usepackage{multicol}
\usepackage{multirow}
\usepackage{booktabs}
\usepackage{siunitx}
\usepackage[hang,flushmargin,symbol]{footmisc}
\usepackage{hyperref}
\usepackage{flushend}
\usepackage{amssymb}
\usepackage{pifont}
\newcommand{\cmark}{\ding{51}}%
\newcommand{\xmark}{\ding{55}}%

\hypersetup{
  colorlinks,%
  citecolor=black,%
  filecolor=black,%
  linkcolor=black,%
  urlcolor=black
}
\newcolumntype{D}{p{2.8cm}}
\newcolumntype{T}{>{\centering\arraybackslash}m{1.2cm}}
\newcolumntype{C}{>{\centering\arraybackslash}m{1.9cm}}
\newcolumntype{L}{>{\centering\arraybackslash}m{2.75cm}}
\newcolumntype{Q}{>{\centering\arraybackslash}m{0.7cm}}
\newcolumntype{Z}{>{\centering\arraybackslash}m{0.5cm}}
\newcolumntype{P}{>{\centering\arraybackslash}m{1.7cm}}

\makeatletter
\if@twocolumn
  \renewcommand\normalsize{%
   \@setfontsize\normalsize\@xpt{12.5pt}%
   \abovedisplayskip=3 mm plus6pt minus 4pt
   \belowdisplayskip=3 mm plus6pt minus 4pt
   \abovedisplayshortskip=0.0 mm plus6pt
   \belowdisplayshortskip=2 mm plus4pt minus 4pt
   \let\@listi\@listI}%

  \renewcommand\small{%
   \@setfontsize\small{8.5pt}\@xpt
   \abovedisplayskip 8.5\p@ \@plus3\p@ \@minus4\p@
   \abovedisplayshortskip \z@ \@plus2\p@
   \belowdisplayshortskip 4\p@ \@plus2\p@ \@minus2\p@
   \def\@listi{\leftmargin\leftmargini
               \parsep 0\p@ \@plus1\p@ \@minus\p@
               \topsep 4\p@ \@plus2\p@ \@minus4\p@
               \itemsep0\p@}%
   \belowdisplayskip \abovedisplayskip}

\else
  \if@smallext
   \renewcommand\normalsize{%
   \@setfontsize\normalsize\@xpt\@xiipt
   \abovedisplayskip=3 mm plus6pt minus 4pt
   \belowdisplayskip=3 mm plus6pt minus 4pt
   \abovedisplayshortskip=0.0 mm plus6pt
   \belowdisplayshortskip=2 mm plus4pt minus 4pt
   \let\@listi\@listI}%

  \renewcommand\small{%
   \@setfontsize\small\@viiipt{9.5pt}%
   \abovedisplayskip 8.5\p@ \@plus3\p@ \@minus4\p@
   \abovedisplayshortskip \z@ \@plus2\p@
   \belowdisplayshortskip 4\p@ \@plus2\p@ \@minus2\p@
   \def\@listi{\leftmargin\leftmargini
               \parsep 0\p@ \@plus1\p@ \@minus\p@
               \topsep 4\p@ \@plus2\p@ \@minus4\p@
               \itemsep0\p@}%
   \belowdisplayskip \abovedisplayskip}
 \else
  \renewcommand\normalsize{%
   \@setfontsize\normalsize{9.5pt}{11.5pt}%
   \abovedisplayskip=3 mm plus6pt minus 4pt
   \belowdisplayskip=3 mm plus6pt minus 4pt
   \abovedisplayshortskip=0.0 mm plus6pt
   \belowdisplayshortskip=2 mm plus4pt minus 4pt
   \let\@listi\@listI}%

  \renewcommand\small{%
   \@setfontsize\small\@viiipt{9.25pt}%
   \abovedisplayskip 8.5\p@ \@plus3\p@ \@minus4\p@
   \abovedisplayshortskip \z@ \@plus2\p@
   \belowdisplayshortskip 4\p@ \@plus2\p@ \@minus2\p@
   \def\@listi{\leftmargin\leftmargini
               \parsep 0\p@ \@plus1\p@ \@minus\p@
               \topsep 4\p@ \@plus2\p@ \@minus4\p@
               \itemsep0\p@}%
   \belowdisplayskip \abovedisplayskip}
  \fi
\let\footnotesize\small
\makeatother

\begin{document}

\title{NeRF and Gaussian Splatting SLAM in the Wild}

\author{Fabian Schmidt$^{1,2}$ \orcidlink{0000-0003-3958-8932},
Markus Enzweiler$^{1}$ \orcidlink{0000-0001-9211-9882}, 
and Abhinav Valada$^{2}$ \orcidlink{0000-0003-4710-3114}

\thanks{$^{1}$ Institute for Intelligent Systems, Esslingen University of Applied Sciences, Germany.}%
\thanks{$^{2}$ Department of Computer Science, University of Freiburg, Germany.}
}



\maketitle

\begin{abstract}
Navigating outdoor environments with visual Simultaneous Localization and Mapping (SLAM) systems poses significant challenges due to dynamic scenes, lighting variations, and seasonal changes, requiring robust solutions. While traditional SLAM methods struggle with adaptability, deep learning-based approaches and emerging neural radiance fields as well as Gaussian Splatting-based SLAM methods, offer promising alternatives. However, these methods have primarily been evaluated in controlled indoor environments with stable conditions, leaving a gap in understanding their performance in unstructured and variable outdoor settings. This study addresses this gap by evaluating these methods in natural outdoor environments, focusing on camera tracking accuracy, robustness to environmental factors, and computational efficiency, highlighting distinct trade-offs. Extensive evaluations demonstrate that neural SLAM methods achieve superior robustness, particularly under challenging conditions such as low light, but at a high computational cost. At the same time, traditional methods perform the best across seasons but are highly sensitive to variations in lighting conditions. The code of the benchmark is publicly available at \url{https://github.com/iis-esslingen/nerf-3dgs-benchmark}.
\end{abstract}


\section{Introduction}

\IEEEPARstart{O}{utdoor} environments pose unique challenges for Simultaneous Localization and Mapping (SLAM) due to their dynamic nature, diverse lighting conditions, and seasonal variations. Therefore, robust SLAM systems are essential for applications such as autonomous navigation and precision agriculture, where accurate and reliable localization is critical. While traditional SLAM methods have been instrumental in enabling autonomous systems to navigate and map environments, their reliance on handcrafted features and discrete representations often limits their adaptability to challenging outdoor domains~\cite{schmidt2024, schmidt2024visual}.

In contrast, deep learning-based approaches improve robustness through advanced feature extraction but face challenges such as dependency on large datasets and limited generalization to unseen scenarios~\cite{vodisch2022continual, valada2018deep, cattaneo2020cmrnet}. 
Emerging representations such as Neural Radiance Fields (NeRF)~\cite{mildenhall2021nerf}, and 3D Gaussian Splatting (3DGS)~\cite{kerbl20233d} offer continuous scene modeling, improved noise handling, and high-resolution reconstructions, addressing limitations in traditional approaches. Despite these advancements, their evaluation has largely focused on indoor environments~\cite{tosi2024nerfs}, leaving a gap in understanding their effectiveness in outdoor settings. Comparative analysis is needed to assess their strengths and limitations, particularly in identifying the most robust and effective scene representation for challenging outdoor domains.

This paper presents a comparative evaluation of traditional SLAM, deep learning-based SLAM, and emerging NeRF- and 3DGS-based approaches in diverse outdoor settings. To this end, we use the ROVER dataset~\cite{schmidt2024rover}, which provides a rich collection of real-world data recorded in various challenging outdoor scenarios. Focusing on key algorithmic components such as pose estimation and scene representation, we analyze the trade-offs in robustness, accuracy, and computational efficiency, providing insights into optimal components for outdoor SLAM. Our findings aim to bridge the gap between theoretical advancements and practical application, guiding future developments in the field of visual SLAM.

\section{Related Work}

SLAM benchmarks are essential for evaluating algorithmic performance in complex and challenging scenarios, serving as a foundation for comparing both traditional and emerging neural SLAM methods. Recent advancements in neural SLAM methodologies, such as NeRF- and 3DGS-based approaches, introduce novel ways for high-quality 3D scene representation and reconstruction. Comprehensive surveys such as~\cite{macario2022comprehensive, tosi2024nerfs} offer an in-depth overview of the algorithmic landscape and its evolution. Well-established traditional benchmarks like EuRoC~\cite{burri2016euroc} and KITTI~\cite{geiger2012kitti} have significantly advanced SLAM evaluation, especially in indoor and urban outdoor settings. The 4Seasons benchmark~\cite{wenzel20214seasons, wenzel20244seasons} further focuses on varying environmental conditions, such as lighting and seasonal changes, within urban outdoor settings. In contrast,~\cite{schmidt2024, schmidt2024visual} emphasize natural, unstructured outdoor scenarios, addressing the challenges of long-term usability in more diverse natural settings.

Neural SLAM benchmarks have emerged to evaluate recent approaches, focusing on NeRF and Gaussian-based scene representations in indoor environments using the Replica~\cite{straub2019replica} and ScanNet~\cite{dai2017scannet} datasets. Xu~\textit{et~al.}~\cite{xu2024perturbations} examined the robustness of RGB-D SLAM systems, both neural and non-neural, under varying noise levels and environmental perturbations, using the Noisy-Replica dataset. The study investigates how different noise models and multi-modal sensor inputs influence SLAM tracking accuracy. Zhou~\textit{et~al.}~\cite{zhou2024evaluating} compared NeRF and 3DGS-based methods for 3D scene reconstruction, tracking, and rendering across Replica and ScanNet datasets. They highlight NeRF's superior photorealistic quality and 3DGS's efficiency, making the latter more suitable for real-time applications. Wang~\textit{et~al.}~\cite{wang2024xrdslam} introduced XRDSLAM, a modular framework for benchmarking SLAM methods, including traditional, NeRF-based and 3DGS-based approaches. Using the Replica dataset, it evaluates tracking, rendering, reconstruction, GPU usage, and FPS to analyze computational and performance trade-offs comprehensively.

Existing neural SLAM benchmarks still face key limitations, such as a lack of diversity in seasonal, weather, and outdoor complexity representation, which restricts the evaluation of SLAM methods for long-term, real-world deployments in dynamic environments. The benchmark introduced in this paper addresses these gaps by focusing on unstructured outdoor scenarios with varying environmental conditions, i.e., season, weather, and lighting, comparing traditional and state-of-the-art NeRF- and 3DGS-based SLAM methods.

\section{Benchmark}

Evaluating SLAM algorithms under diverse conditions is critical for understanding their real-world performance. The following sections detail the evaluated algorithms, describe the evaluation methodology, and present experiments along with the analysis of the results, focusing on accuracy, robustness, and computational efficiency.

\subsection{SLAM Methods}
The benchmark includes diverse SLAM methods, spanning traditional, deep learning-based, NeRF-based, and 3DGS-based approaches. The different algorithmic components of these methods, such as pose estimation techniques, scene encoding strategies, geometry representations, and the ability to handle loop closures, are summarized in Table~\ref{table:slam_methods}. 

\begin{table}
    \centering
    \footnotesize
    \caption{Overview of the SLAM methods and their components, where F-F denotes Frame-To-Frame and F-M denotes Frame-To-Model.}
    \setlength{\tabcolsep}{2pt}
        \begin{tabular*}{\linewidth}{@{\extracolsep{\fill}} lcccc} 
            \toprule
            \multirow{ 2}{*}{\textbf{SLAM}} & \textbf{Pose} & \textbf{Scene} & \textbf{Geometry} & \textbf{Loop} \\
            ~ & \textbf{Estimation} & \textbf{Encoding} & \textbf{Repr.} & \textbf{Closure} \\
            \toprule
            ORB-SLAM3 & F-F & n/a & n/a & \cmark \\
            DROID-SLAM & F-F & n/a & n/a & \xmark \\
            DPV-SLAM & F-F & n/a & n/a & \cmark \\
            \midrule
            ORBEEZ-SLAM & F-F (ORB2) & Hash Grid & Density & \xmark \\ 
            GO-SLAM & F-F (DROID) & Hash Grid & SDF & \xmark\\ 
            GlORIE-SLAM & F-F (DROID) & Neural Points & Occupancy & \cmark \\ 
            Co-SLAM & F-M & Hash Grid & SDF & \xmark\\
            \midrule
            MonoGS & F-M & 3D Gaussians & Density & \xmark \\ 
            Photo-SLAM & F-F (ORB3) & 3D Gaussians & Density & \cmark \\ 
            Splat-SLAM & F-F (DROID) & 3D Gaussians & Density & \xmark \\
            GS-ICP-SLAM & F-F (G-ICP) & 3D Gaussians & Density & \xmark \\
            \bottomrule
        \end{tabular*}%
    \label{table:slam_methods}
\end{table}

Traditional methods, such as ORB-SLAM3~\cite{campos2021orb-slam3}, serve as a baseline with well-established feature-based techniques. The evaluated deep learning-based methods, including DROID-SLAM~\cite{teed2021droid-slam} and DPV-SLAM~\cite{lipson2024dpv-slam}, leverage neural end-to-end architectures for pose estimation. NeRF-based methods evaluated in our experiments, including Orbeez-SLAM~\cite{chung2023orbeez}, GlORIE-SLAM~\cite{zhang2024glorie}, Co-SLAM~\cite{wang2023coslam}, and GO-SLAM~\cite{zhang2023goslam}, utilize neural radiance fields for photorealistic scene representation and localization. These methods leverage advanced scene encoding techniques such as Multi-Layer Perceptrons~(MLPs) combined with Hash Grids or Neural points to encode dense, continuous volumetric representations. In contrast, 3DGS-based methods including MonoGS~\cite{matsuki2024monogs}, Photo-SLAM~\cite{huang2024photoslam}, Splat-SLAM~\cite{sandstrom2024splatslam}, and GS-ICP-SLAM~\cite{ha2024gs-icp-slam}, focus on efficient 3D Gaussian-based scene representations. These methods excel in balancing computational efficiency with real-time performance and precise geometric reconstruction. Both NeRF- and 3DGS-based methods rely on either Frame-to-Frame (F-F) pose estimation using external trackers or Frame-to-Model (F-M) approaches.

\begin{figure}
\centering
\footnotesize
  \setlength{\tabcolsep}{0.0cm}
  {\renewcommand{\arraystretch}{1}
    \begin{tabular}{p{\linewidth}}
        \includegraphics[width=\linewidth]{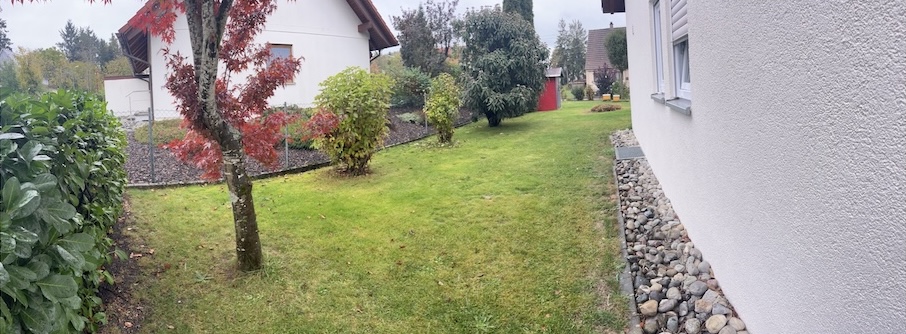} \\[0.5em]
        \multicolumn{1}{c}{(a) Garden Small location} \\[0.5em]
        
        \includegraphics[width=\linewidth]{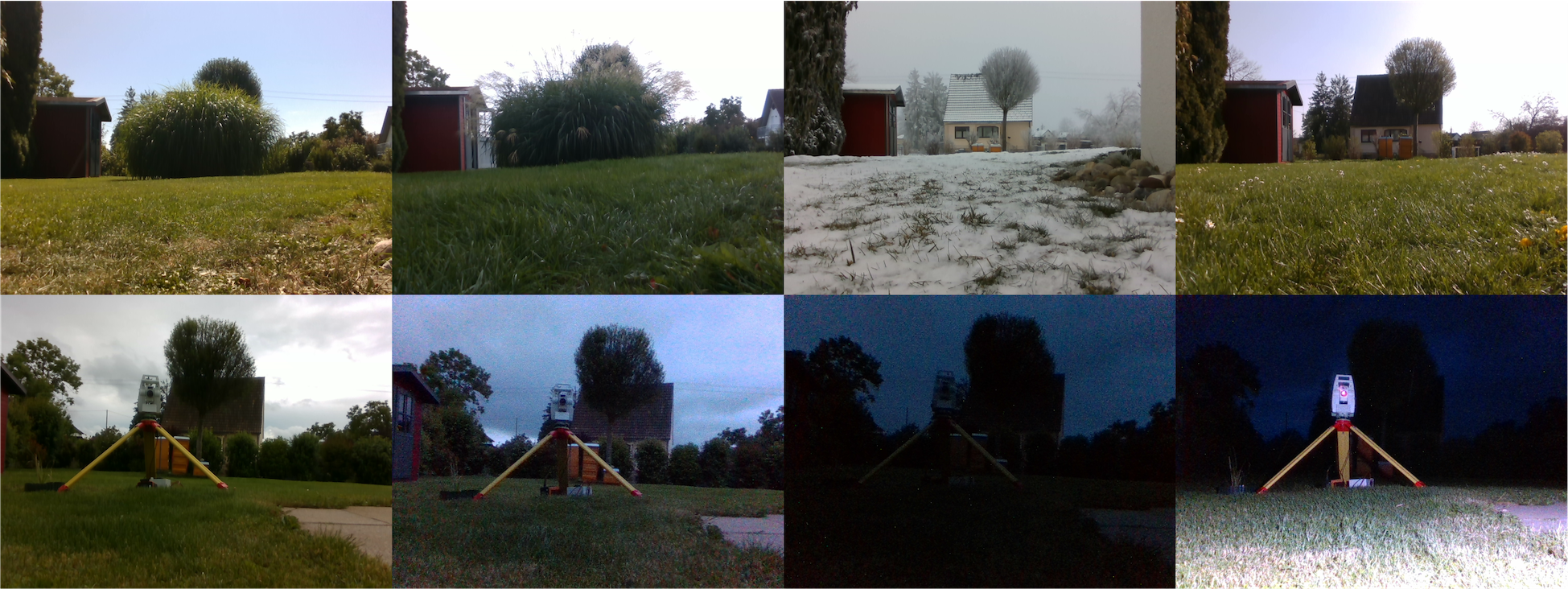} \\[0.5em]
        \multicolumn{1}{c}{(b) Environmental variations} \\
    \end{tabular}}
    \caption{Illustration of the ROVER dataset sequences from Garden Small location (a), captured across various environmental conditions (b), where the top row depicts the different seasons (summer, autumn, winter, spring), while the bottom row shows the varying lighting conditions (daylight, dusk, night, night with external lighting).}
\label{fig:dataset_sequences}
\end{figure}

\subsection{Evaluation Methodology}
For evaluations, we use the ROVER dataset~\cite{schmidt2024rover}, specifically focusing on the \textit{Garden Small} location, which combines dense vegetation with textureless surfaces, illustrated in Figure~\ref{fig:dataset_sequences}, increasing the difficulty of maintaining consistent camera tracking. This subset includes eight sequences representing various environmental conditions, covering all seasons and four lighting scenarios (day, dusk, night, and night with external lighting), as well as diverse weather conditions, including sunny, cloudy, windy, and snow scenarios. Due to GPU RAM limitations, we reduced each sequence from its original three rounds to a single round while still enabling loop closure. We assessed the accuracy of camera tracking for the evaluated SLAM methods using the Root Mean Squared Error (RMSE) of the Absolute Trajectory Error (ATE) as the primary metric, following the methodology described in~\cite{schmidt2024visual}. In addition, we monitored GPU utilization and runtime to evaluate the computational efficiency of the methods, including any optimizations applied during post-processing. We used open-source implementations of the selected SLAM algorithms, running them with default parameters and pretrained weights from their original setups without any additional optimization. We conducted the experiments on an AMD Ryzen 9 7950X3D CPU with 64 GB RAM and an NVIDIA RTX A6000 GPU, running each SLAM method in its own dedicated Docker container.

\subsection{Experiments and Results}
We present the evaluation of SLAM methods across diverse environmental conditions, focusing on the natural outdoor domains. The experiments analyze the camera tracking performance of the SLAM algorithms under different seasons and four distinct lighting conditions. Additionally, we monitor GPU utilization alongside process runtimes on the spring sequence. 

\begin{table}
    \centering
    \footnotesize
    \caption{Camera tracking RMSE ATE [\SI{}{\meter}] across different seasons, with the best results highlighted in bold. A '-' denotes method failure during execution, an 'x' indicates results exceeding \SI{100}{\meter}, and a '*' signifies that the mean results are not directly comparable due to missing data.}
    \setlength{\tabcolsep}{1.05pt}
        \begin{tabular*}{1\linewidth}{@{\extracolsep{\fill}} clccccr} 
            \toprule
            & \textbf{SLAM} & \textbf{Summer} & \textbf{Autumn} & \textbf{Winter} & \textbf{Spring} & \textbf{Mean $\pm$ Std} \\ \midrule
            \multirow{8}{*}{\rotatebox{90}{Mono}} & ORB-SLAM3 & 4.28 & 4.49 & 2.89 & 5.25 & 4.23 $\pm$ 0.98 \\
            & DROID-SLAM & 2.39 & 1.36 & \textbf{1.73} & 2.83 & \textbf{2.08} $\pm$ 0.66 \\
            & DPV-SLAM & 2.59 & 1.59 & 2.02 & 2.20 & 2.10 $\pm$ 0.42 \\
            & GO-SLAM & \textbf{1.74} & \textbf{1.07} & 3.63 & 2.01 & 2.11 $\pm$ 1.09 \\
            & ORBEEZ-SLAM & - & - & - & - & \multicolumn{1}{c}{\; -} \\
            & MonoGS & - & 3.08 & - & 3.36 & 3.22* $\pm$ 0.20 \\
            & Photo-SLAM & 4.48 & 5.05 & 3.46 & 7.19 & 5.05 $\pm$ 1.57 \\
            & Splat-SLAM & 2.99 & 1.26 & 2.27 & \textbf{1.96} & 2.12 $\pm$ 0.65 \\
            \midrule
            \multirow{10}{*}{\rotatebox{90}{RGB-D}} & ORB-SLAM3 & \textbf{0.50} & \textbf{0.65} & 0.56 & \textbf{0.58} & \textbf{0.57} $\pm$ 0.06 \\
            & DROID-SLAM & 0.91 & 1.99 & \textbf{0.55} & 0.71 & 1.04 $\pm$ 0.65 \\
            & GO-SLAM & 0.56 & 0.68 & 0.56 & 0.71 & 0.63 $\pm$ 0.08 \\
            & ORBEEZ-SLAM & 0.58 & 1.07 & 0.59 & 0.69 & 0.73 $\pm$ 0.23 \\
            & Co-SLAM & x & x & x & x & \multicolumn{1}{c}{\; x} \\
            & GlORIE-SLAM & 3.34 & 2.34 & 3.84 & 2.82 & 3.08 $\pm$ 0.65 \\
            & MonoGS & 0.61 & 1.48 & 0.69 & 0.81 & 0.90 $\pm$ 0.40 \\
            & Photo-SLAM & 0.52 & 0.88 & 0.56 & 0.63 & 0.65 $\pm$ 0.16 \\
            & GS-ICP-SLAM & 2.42 & 2.22 & 2.12 & 3.43 & 2.55 $\pm$ 0.60 \\
            \bottomrule
        \end{tabular*}
    \label{table:results_seasons}
\end{table}

\begin{table}
    \centering
    \footnotesize
    \caption{Camera tracking RMSE ATE [\SI{}{\meter}] for different lighting conditions.}
    \setlength{\tabcolsep}{1.25pt}
        \begin{tabular*}{1\linewidth}{@{\extracolsep{\fill}} clccccr} 
            \toprule
            & \textbf{SLAM} & \textbf{Day} & \textbf{Dusk} & \textbf{Night} & \textbf{+ Light} & \textbf{Mean / Std} \\
            \midrule
            \multirow{8}{*}{\rotatebox{90}{Mono}} & ORB-SLAM3 & 5.00 & - & - & 4.63 & 4.82* $\pm$ 0.13 \\
            & DROID-SLAM & 2.21 & 4.33 & 4.56 & 3.71 & 3.70 $\pm$ 0.44 \\
            & DPV-SLAM & 2.65 & 4.15 & 5.02 & 4.06 & 3.97 $\pm$ 0.49 \\
            & GO-SLAM & \textbf{1.27} & 4.55 & \textbf{3.65} & 3.80 & \textbf{3.32} $\pm$ 0.52 \\
            & ORBEEZ-SLAM & - & - & - & - & \multicolumn{1}{c}{\; -} \\
            & MonoGS & - & - & - & \textbf{3.67} & 3.67* $\pm$ 0.00 \\
            & Photo-SLAM & 5.16 & 4.85 & - & 4.89 & 4.97* $\pm$ 0.06 \\
            & Splat-SLAM & 1.28 & \textbf{2.94} & - & 4.24 & 2.82* $\pm$ 0.81 \\
            \midrule
            \multirow{10}{*}{\rotatebox{90}{RGB-D}} & ORB-SLAM3 & 0.59 & 5.32 & 4.99 & 0.72 & 2.91 $\pm$ 2.13 \\
            & DROID-SLAM & 0.57 & 1.21 & 3.52 & 0.93 & 1.56 $\pm$ 1.17 \\
            & GO-SLAM & \textbf{0.56} & 1.17 & 3.16 & 0.83 & \textbf{1.43} $\pm$ 1.04 \\
            & ORBEEZ-SLAM & 0.65 & - & - & 0.81 & 0.73* $\pm$ 0.06 \\
            & Co-SLAM & x & x & x & x & \multicolumn{1}{c}{\; x} \\
            & GlORIE-SLAM & 3.63 & 2.54 & - & 9.33 & 5.17* $\pm$ 3.42 \\
            & MonoGS & 1.00 & \textbf{0.86} & \textbf{1.26} & 2.92 & 1.51 $\pm$ 0.89 \\
            & Photo-SLAM & 0.59 & 5.94 & - & \textbf{0.59} & 2.37* $\pm$ 2.73 \\
            & GS-ICP-SLAM & 3.78 & 2.97 & 3.58 & 3.34 & 3.42 $\pm$ 0.26 \\
            \bottomrule
        \end{tabular*}
    \label{table:results_lighting}
\end{table}

We observe that different seasons and changing environmental appearance only have just a minor impact on the camera tracking performance, as shown in Table~\ref{table:results_seasons}. For monocular methods, DROID-SLAM emerged as the best-performing algorithm, achieving a mean ATE of \SI{2.08}{\meter} with consistent results across all seasons and minimal outliers, primarily in spring. It demonstrates strong robustness, performing reliably across different environmental conditions. DPV-SLAM, GO-SLAM, and Splat-SLAM showed competitive performance, while ORB-SLAM3 and Photo-SLAM struggled significantly. Methods such as MonoGS and ORBEEZ-SLAM were either only partially functional or failed entirely. For RGB-D methods, ORB-SLAM3 provided the most accurate results, with a mean ATE of \SI{0.57}{\meter} and remarkable robustness (standard deviation of \SI{0.06}{\meter}) across all seasons. GO-SLAM, Photo-SLAM, and ORBEEZ-SLAM performed similarly well, maintaining reliable performance. MonoGS and DROID-SLAM showed moderate performance, generally performing well but exhibiting higher variability in autumn, where environmental conditions significantly impacted their mean ATE. In contrast, methods such as GS-ICP-SLAM and GlORIE-SLAM produced consistently poor results, while Co-SLAM frequently diverged, rendering them ineffective. 

Table~\ref{table:results_lighting} demonstrates the significant impact of lighting conditions on monocular methods. GO-SLAM was the most robust monocular method, with consistent performance across all lighting conditions but with a higher mean ATE of \SI{3.32}{\meter}. Other methods, including DROID-SLAM and DPV-SLAM, showed moderate performance but also struggled in low-light settings. ORB-SLAM3 and Photo-SLAM were highly sensitive to darker scenarios, while ORBEEZ-SLAM was non-functional in any lighting scenario.
RGB-D methods demonstrated better adaptability to lighting variations. GO-SLAM and DROID-SLAM performed well in most scenarios but were significantly impacted by complete darkness. MonoGS performed effectively in darker environments but struggled under artificial lighting. Photo-SLAM, ORB-SLAM3, and ORBEEZ-SLAM provided reliable results in brighter scenarios, but their performance degraded significantly in darker conditions, often failing to produce any results. In contrast, GS-ICP-SLAM and GlORIE-SLAM showed consistently poor results with higher errors, while Co-SLAM again frequently diverged and failed entirely.

\begin{figure*}
  \centering
  \footnotesize
  \setlength{\tabcolsep}{0.05cm}
  {\renewcommand{\arraystretch}{1}
    \begin{tabular*}{\textwidth}{@{}l@{\extracolsep{\fill}}r@{}}
        \includegraphics[height=5.35cm]{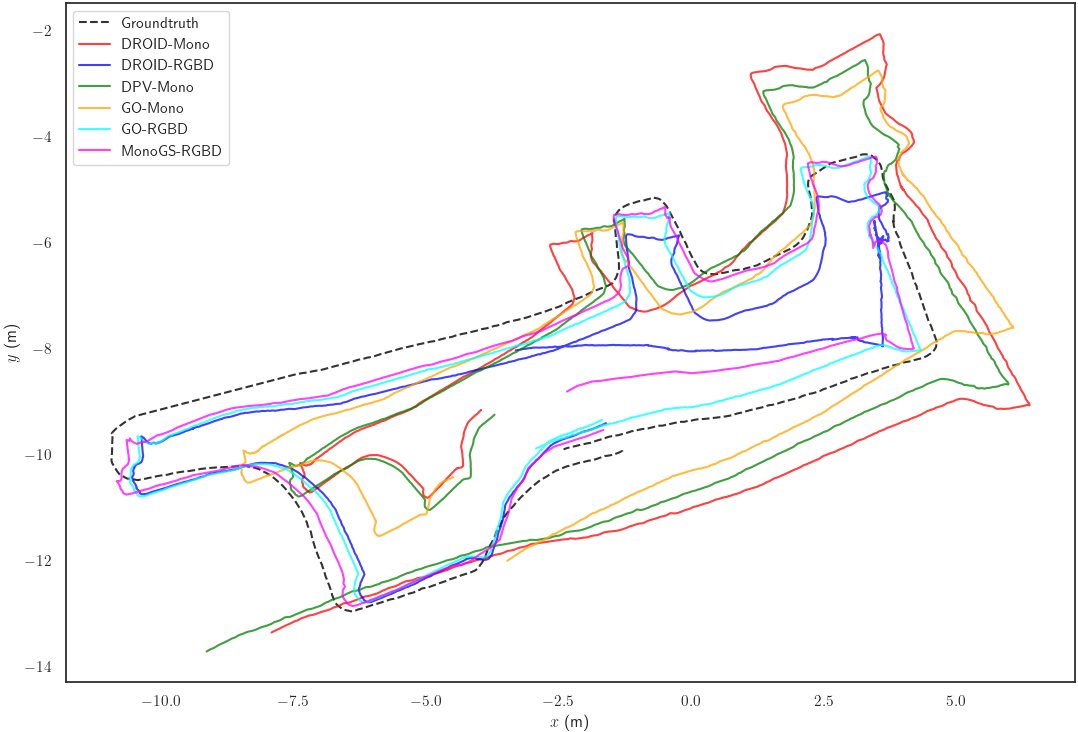} &
        \includegraphics[height=5.35cm]{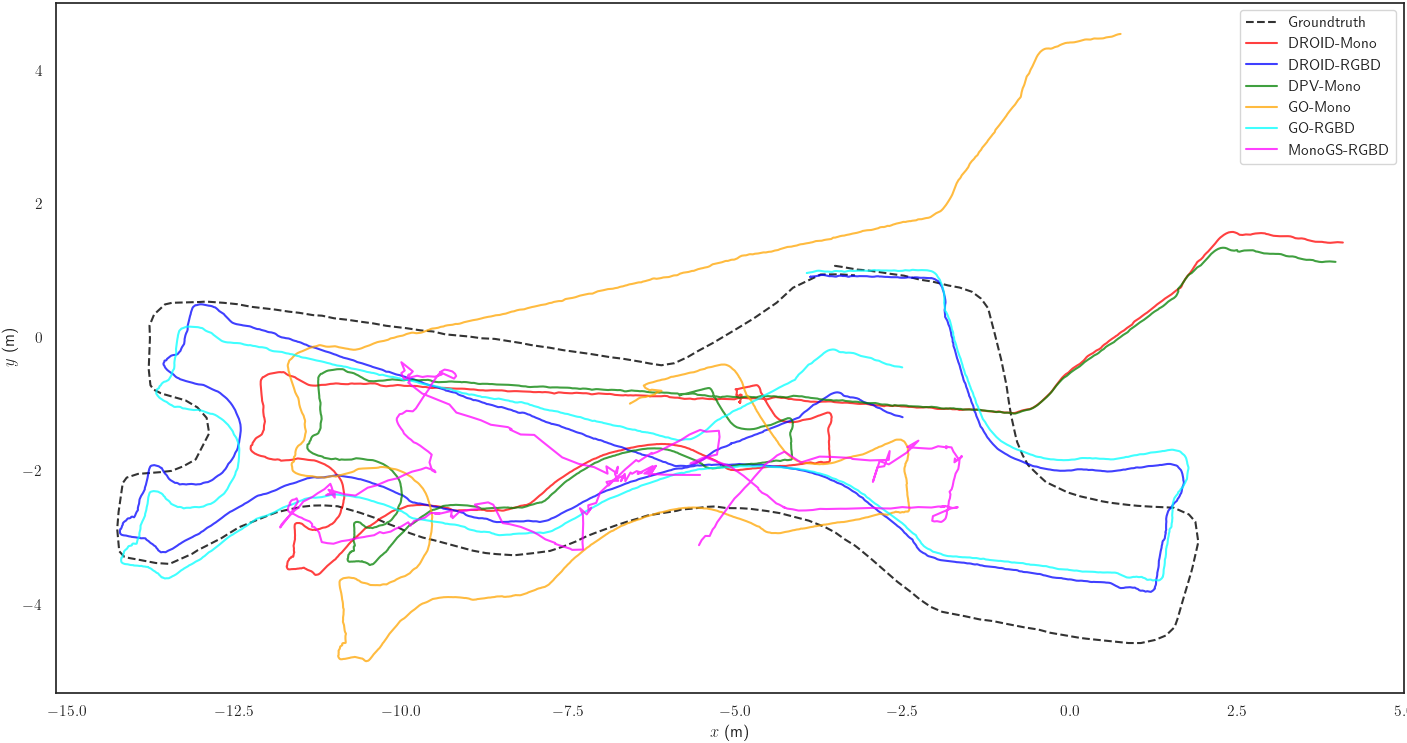} \\
        \multicolumn{1}{c}{(a) Summer} & \multicolumn{1}{c}{(b) Night + Light} \\[0.5em]
    \end{tabular*}}
    \caption{Qualitative trajectory comparisons for the top three Mono (GO-SLAM, DROID-SLAM, DPV-SLAM) and RGB-D (GO-SLAM, MonoGS, DROID-SLAM) SLAM methods in two scenarios, highlighting differences between the modalities. (a) Summer: Mono methods struggle with scale ambiguity and drift, while RGB-D methods also face drift problems, with DROID-SLAM and MonoGS being notably affected, whereas GO-SLAM achieves the best overall performance. (b) Night + Light: Mono methods face severe scaling problems, with GO-SLAM additionally suffering significant drift. For RGB-D, DROID-SLAM and GO-SLAM exhibit noticeable drift, while MonoGS struggles with consistent tracking and scaling.}
    \label{fig:qual_res_trajectories}
\end{figure*}

RGB-D configurations significantly outperform their monocular counterparts in overall camera tracking performance and robustness, particularly under diverse environmental conditions, as shown in Figure~\ref{fig:qual_res_trajectories}. While monocular methods struggle with scale ambiguity and drift, RGB-D approaches handle drift more effectively, though they still face issues in challenging conditions like complete darkness. GO-SLAM in RGB-D mode demonstrated the most robust and accurate performance across various scenarios, albeit with substantial GPU utilization. MonoGS and DROID-SLAM in RGB-D mode also show competitive performance, demonstrating their ability to match traditional methods like ORB-SLAM3, which struggles under difficult lighting. However, MonoGS requires high GPU resources and is not real-time capable, while DROID-SLAM, despite its high GPU demand, offers faster performance. ORBEEZ-SLAM and Photo-SLAM, although more sensitive to darker scenarios, provide strong results across seasons with significantly lower GPU utilization and faster runtimes, offering an efficient solution for computationally constrained environments.

\begin{table}
    \centering
    \caption{GPU utilization and runtime for different SLAM methods, monitored on the spring sequence, having a total duration of \SI{2}{\minute} and \SI{18}{\second} at 30 FPS.}
    \fontsize{8pt}{8pt}\selectfont
    \begin{tabular*}{1\linewidth}{@{\extracolsep{\fill}}clccr}
        \toprule
        & \textbf{SLAM} & \multicolumn{2}{c}{\textbf{GPU Memory [GB]}} & \textbf{Runtime} \\ \cmidrule{3-4}
        & & Median & Max &  \\
        \midrule
        \multirow{7}{*}{\rotatebox{90}{Mono}} & DROID-SLAM & 14.16 & 16.64 & 5 \SI{}{\minute} 29 \SI{}{\second} \\
        & DPV-SLAM & 2.32 & 2.96 & 2 \SI{}{\minute} 13 \SI{}{\second} \\
        & GO-SLAM & 12.37 & 14.33 & 8 \SI{}{\minute} 47 \SI{}{\second} \\
        & MonoGS & 3.48 & 4.58 & 14 \SI{}{\minute} 52 \SI{}{\second} \\
        & Photo-SLAM & 3.04 & 8.61 & 2 \SI{}{\minute} 56 \SI{}{\second} \\
        & Splat-SLAM & 18.79 & 23.54 & 10 \SI{}{\minute} 18 \SI{}{\second} \\ \midrule
        & DROID-SLAM & 14.15 & 16.77 & 5 \SI{}{\minute} 36 \SI{}{\second} \\ 
        \multirow{7}{*}{\rotatebox{90}{RGB-D}} & GO-SLAM & 12.55 & 14.29 & 9 \SI{}{\minute} 00 \SI{}{\second} \\
        & ORBEEZ-SLAM & 4.34 & 7.44 & 4 \SI{}{\minute} 14 \SI{}{\second} \\
        & Co-SLAM & 4.62 & 4.63 & 10 \SI{}{\minute} 42 \SI{}{\second} \\
        & GlORIE-SLAM & 11.39 & 14.94 & 8 \SI{}{\minute} 22 \SI{}{\second} \\
        & MonoGS & 13.20 & 14.65 & 55 \SI{}{\minute} 47 \SI{}{\second} \\
        & Photo-SLAM & 3.06 & 8.88 & 3 \SI{}{\minute} 02 \SI{}{\second} \\
        & GS-ICP-SLAM & 15.26 & 20.37 & 10 \SI{}{\minute} 44 \SI{}{\second} \\
        \bottomrule
    \end{tabular*}
    \label{table:gpu_runtime}
    \vspace{-0.3cm}
\end{table}

The evaluations reveal insights into the algorithmic components of evaluated SLAM methods. Among deep learning-based approaches, DROID-SLAM achieved high accuracy at the cost of substantial GPU usage (refer to Table~\ref{table:gpu_runtime}), while DPV-SLAM provided a lightweight, real-time alternative with comparable results. In NeRF-based methods, the mapping properties of GO-SLAM, leveraging hash grids and signed distance functions, consistently outperformed GlORIE-SLAM’s neural points and occupancy grid in both accuracy and resource efficiency. GO-SLAM’s frame-to-frame pose estimation was also superior to frame-to-model approaches such as Co-SLAM, which failed to perform reliably. External trackers significantly influenced performance; ORBEEZ-SLAM, relying on ORB2-SLAM2~\cite{mur2017orb2}, matched GO-SLAM’s accuracy when functional but struggled under darker conditions. In contrast, GO-SLAM’s use of DROID-SLAM as an external tracker ensured reliable tracking at the expense of increased GPU usage.
In 3DGS methods, Splat-SLAM's frame-to-frame pose estimation outperformed MonoGS's frame-to-model approach in monocular setups. In RGB-D mode, MonoGS delivered significantly more stable results than Photo-SLAM. Splat-SLAM, using DROID-SLAM as an external tracker, outperformed Photo-SLAM, relying on ORB-SLAM3 but demanded significantly more GPU resources.

\section{Conclusion}
In this study, we evaluated a wide range of SLAM algorithms under diverse outdoor conditions, demonstrating the impact of seasonality and lighting on performance. While seasonal variations had a minor effect, lighting conditions, especially in low-light scenarios, had a significant impact, particularly for monocular SLAM methods. GO-SLAM, which incorporates DROID-SLAM as an external tracker, provides the best camera tracking performance and robustness in both monocular and RGB-D modes but demands high CPU usage. DROID-SLAM offers a strong monocular and RGB-D alternative that is sensitive to darker scenarios and requires high GPU usage, while DPV-SLAM is a lightweight, real-time monocular option with comparable results. Photo-SLAM and ORBEEZ-SLAM in RGB-D mode deliver efficient performance with lower resource demands but are sensitive to low-light conditions, similar to traditional methods such as ORB-SLAM3. The results underscore the trade-offs between performance, robustness, and computational efficiency across different SLAM methods, providing valuable insights for selecting the most suitable approach based on environmental conditions and resource constraints.



\footnotesize
\bibliographystyle{IEEEtran}
\bibliography{bibliography}

\vspace{11pt}

\vfill

\end{document}